\documentclass{article}

 \usepackage[sglblindworkshop,final]{neurips_2025}
 \usepackage{amsmath}
 \usepackage{amssymb}
 \usepackage{bm}
 \usepackage{graphicx}
 \usepackage{caption}
\usepackage{subcaption}
\usepackage{hyperref}

\usepackage[utf8]{inputenc} 
\usepackage[T1]{fontenc}    
\usepackage{hyperref}       
\usepackage{url}            
\usepackage{booktabs}       
\usepackage{amsfonts}       
\usepackage{nicefrac}       
\usepackage{microtype}      
\usepackage{xcolor}         

\usepackage{soul} 

\newcommand{\commentout}[1]{}

\title{Capturing non-Markovian dynamics in non-equilibrium stochastic systems using flow matching}

\author{%
  Bhargav Sriram Siddani \\
  Lawrence Berkeley National Laboratory\\
  \texttt{bsiddani@lbl.gov} \\
  \And
  John B. Bell \\
  Lawrence Berkeley National Laboratory\\
  \texttt{jbbell@lbl.gov} \\
  \And
  Alejandro L. Garcia \\
  San Jose State University\\
  \texttt{alejandro.garcia@sjsu.edu} \\
  \And
  Ishan Srivastava \\
  Lawrence Berkeley National Laboratory\\
  \texttt{isriva@lbl.gov} \\
}

\begin{document}
\maketitle

\begin{abstract}
Hydrodynamic models of stochastic particle systems represented by coarse-grained stochastic partial differential equations (SPDE), such as the regularized Dean-Kawasaki (DK) equation, do not accurately capture the short-time system dynamics that is dominated by non-Markovian effects, and low particle density regimes where the distributions are highly non-Gaussian.
We develop a generative flow matching method that directly models the probability distribution of fluxes from particle simulations that explicitly incorporates non-Markovian and non-Gaussian effects.
As a demonstration, we use this method to simulate the Kramers first passage time problem for a system of non-interacting Brownian particles. We show the model accurately captures the short-time behavior and provides better predictions of the statistical moments of the number density when compared against the solution of the Markovian baseline, regularized DK equation.
\end{abstract}

\section{Introduction}
A wide range of phenomena in areas as diverse as fluid dynamics, chemistry, biology and the social sciences can be described by stochastic particle systems.  For small to modest sized systems, the dynamics can be modeled directly using particle-based simulations, such as molecular dynamics. However, for very large systems, direct simulation of particle dynamics becomes prohibitively expensive.  One approach to reducing the computational cost of many-particle system simulations, originally introduced by Dean \cite{dean_langevin_1996} and Kawasaki \cite{kawasaki1994stochastic}, is to coarse-grain the particle dynamics to obtain a stochastic partial differential equation (SPDE) for the number density, similar to the formulation of the dynamic density functional theory.  Development and analysis of these types of coarse-grained SPDE models has been an active area of research for the past several decades.  See~\cite{illien2025dean,te2023perspective} for recent reviews of work in this area.

Here, we consider a representative system of $N$ non-interacting Brownian particles under the influence of an external potential, $V(x)$. In the overdamped limit, the dynamics of the position, $\bm{X}_{i}(t)$, of the $i^{th}$ particle is given by the stochastic ODE
\begin{equation}
    d\bm{X}_{i} = -\mu \nabla V(\bm{X}_{i}) dt + d \bm{B}_{t}^{i}, \, i=1,2,...,N, 
\end{equation}
where $\mu$ is a Stokes mobility and the $\bm{B}_t$ are independent Brownian motions.
An SPDE that describes the evolution of the particles can be formally derived using It\^o's formula~\cite{dean_langevin_1996} but the resulting system, which includes the divergence of space-time white noise, is highly irregular.  Typically, a coarse-grained regularized version of the SPDE is introduced by incorporating a high-frequency cutoff to the white noise. This results in a hydrodynamic equation for number density, $q$, of the form
\begin{equation}\label{eq:number_density_DK}
    dq = \nabla \cdot \frac{1}{2} \nabla q \, dt + \nabla \cdot \sqrt{q}\, d\bm{W}^{M} \, + \nabla \cdot \left( \mu \, q \nabla V\right) dt  \, 
    ,
\end{equation}
where  $d\bm{W}^{M}$ represents the regularized noise.
Here we regularize the noise by discretizing Eq. (\ref{eq:number_density_DK}) on a one-dimensional finite volume mesh using an Euler-Maruyama discretization that gives
\begin{equation}
    \frac{q_j^{n+1} - q_j^n}{\Delta t} = - \frac {( \bar{\bm{F}}^B_{j+1/2}-\bar{\bm{F}}^B_{j-1/2} )}{\Delta x}
     - \frac {( \tilde{\bm{F}}^B_{j+1/2}-\tilde{\bm{F}}^B_{j-1/2} )}{\Delta x}
      - \frac {( {\bm{F}}^V_{j+1/2}-{\bm{F}}^V_{j-1/2} )}{\Delta x}
\end{equation}
where the numerical fluxes are given by
\[
\bar{\bm{F}}^B_{j+1/2} = - \frac{q_{j+1}^n - q_{j}^n}{2 \Delta x} ,\; 
\tilde{\bm{F}}^B_{j+1/2} = - A(q_j^n ,q_{j+1}^n)\frac{Z_{j+1/2}^n}{\sqrt{\Delta x \Delta t}}, \,
{\bm{F}}^V_{j+1/2} = -\frac{q_j^n + q_{j+1}^n}{2} \mu V'(x_{j+1/2}).
\]\
Here $Z_{j+1/2}^n$ are normally distributed random numbers, and
\[
A(q_1,q_2) = \sqrt{\frac{\mathrm{max}(q_1,0) + \mathrm{max}(q_2,0)}{2}}
\]
approximates $\sqrt{q}$ at cells faces while avoiding numerical issues if $q$ becomes negative. Further details regarding the regularization and finite-volume discretization are discussed in \cite{djurdjevac2025hybrid}. 

The regularized DK equation has been shown to correctly predict long-time system statistics, compared to particle simulations, when the number of particles is sufficiently large in every finite-volume cell. However, there are significant differences in statistics when there are regions with low particle counts~\cite{djurdjevac2025hybrid}. Moreover, it is now well-known that the dynamics of such diffusive, stochastic systems are significantly influenced by memory effects where the initial conditions impact the evolution of the system trajectories at early times~\cite{derrida2009current,leibovich2013everlasting,banerjee2022role}.  Classical Mori-Zwanzig theory from statistical physics predicts coarse-graining to introduce a memory term into the dynamics (cf. \cite{grabert2006projection}). In the present context, the memory reflects the sensitivity of the coarse-graining of the particles dynamics to initial conditions, which is lost in the regularized coarse-grained DK setting.

\section{Background}
\label{background}
In the present work, we alleviate the shortcomings of the regularized DK equation through data-driven learning for the Brownian diffusion flux, $\bm{F}^{B} \equiv \bar{\bm{F}}^{B} + \tilde{\bm{F}}^{B}$. Specifically we investigate the impact of including memory on the predictive capability of the machine learning (ML) model.
Recent studies have used machine learning to model stochastic processes in dynamical systems. For example, diffusion models were used to learn Lagrangian trajectories of particles in turbulence \cite{li2024synthetic,li2024generative_ijmf}, and a statistics-informed neural network (SINN) framework~\cite{zhu2023learning} was developed to learn non-Markovian stochastic dynamics using recurrent neural networks trained with statistics-based loss functions. Furthermore, machine learning methods have been used to learn and parameterize the memory kernels of non-Markovian dynamics in the context of generalized Langevin equations~\cite{xie2024ab,she2023data,bassi2024learning}.
\section{Method}
\label{method}
We define the current, $J(\bm{x}, t)$, as the net number of particles crossing a face between two adjacent computational cells from time step $t$ to step $t+\Delta t$, and which is related to the flux $\bm{F}$ as 
$J = \left(\bm{F} \cdot \bm{A}\right) \Delta t$, where $\bm{A}$ is the face area.
The probability density function (PDF) of $J$ depends on the number of particles in the cells to the left and right of the face, $N_{L}$ and $N_{R}$ respectively, and exhibits non-Gaussian properties when either value is low~\cite{djurdjevac2025hybrid}. Furthermore, the current $J(\bm{x}, t)$ is also non-Markovian, with probability distribution that depends on the history of the system~\cite{derrida2009current,leibovich2013everlasting,banerjee2022role}.
However, the stochastic flux, $\tilde{\bm{F}}^{B}$, used in the regularized DK equation (Eq.~\ref{eq:number_density_DK}) is a Gaussian distribution that is uncorrelated in space and time. Therefore, we use generative modeling based on the flow matching (FM) framework~\cite{lipman2023flow,lipman2024flowmatchingguidecode,ma2024SiT} to approximate the non-Markovian and non-Gaussian $J(\bm{x}, t)$ distribution from particle simulations to accurately predict short-time dynamics.

\paragraph{Flow matching overview} Flow matching builds a probability path $p_{\tau}, \, 0 \leq \tau \leq 1,$ from a known source distribution $p_{0} = s$ to the target distribution $p_{1} = r$. The goal of flow matching is to learn the parameters $\theta$ of a velocity field neural network, $u_{\tau}^{\theta}$. Once trained, the velocity field $u_{\tau}^{\theta}$ can be used to convert a sample from the source distribution $s$ to a sample from the target distribution $r$ by solving the ordinary differential equation (ODE) from $\tau=0$ to $\tau = 1$. The neural network $u_{\tau}^{\theta}$ is trained using the conditional flow matching loss,
\begin{equation}
    \mathcal{L}_{\text{CFM}}(\theta) = \mathbb{E}_{\tau, Z_{\tau}, Z_{1}} ||u_{\tau}^{\theta}(Z_{\tau}) - u_{\tau}(Z_{\tau}|Z_{1}) ||^{2},
\end{equation}
where $\tau \sim U[0,1]$ (uniform distribution), $Z_{0} \sim s$, $Z_{1} \sim r$, $Z_{\tau} = a_{\tau} Z_{0} + b_{\tau} Z_{1}$, and $u_{\tau}(Z_{\tau}|Z_{1}) = \dot{a}_{\tau} Z_{0} + \dot{b}_{\tau} Z_{1}$ is called conditional velocity field.

\paragraph{Problem specific details}
We use flow matching to learn the PDF of $J(\bm{x}, t)$ conditioned upon the history of the local state of the system from $k$ previous time steps, i.e., $(\bm{N}_{L}^{k}, \bm{N}_{R}^{k}, \bm{J}^{k})$, where
\begin{eqnarray*}
\bm{N}_{L}^{k} = \{N_{L}(\bm{x}, t), N_{L}(\bm{x}, t-\Delta t), ...,N_{L}(\bm{x}, t-k\Delta t)\}, \\ 
\bm{N}_{R}^{k} = \{N_{R}(\bm{x}, t), N_{R}(\bm{x}, t-\Delta t), ...,N_{R}(\bm{x}, t-k\Delta t)\}, \\
\bm{J}^{k} = \{J(\bm{x}, t-\Delta t), J(\bm{x}, t-2\Delta t), ..., J(\bm{x}, t-k\Delta t)\}.
\end{eqnarray*}
Therefore, $u_{\tau}^{\theta}$ takes $(\bm{N}_{L}^{k}, \bm{N}_{R}^{k}, \bm{J}^{k})$ as additional inputs to capture the non-Markovian effects of $J(\bm{x},t)$. To capture the large tails of the target distribution, we chose the source distribution $s$ to be the Student's t-distribution with a zero mean, unity scale, and a value of 4 for degrees of freedom, $a_{\tau} = \cos{(\frac{1}{2}\pi\tau)}$ and $b_{\tau} = \sin{(\frac{1}{2}\pi\tau)}$. Each mini-batch of training samples $Z_{1}$ was generated from a one-dimensional, equilibrium distribution of Brownian walkers across 186 computational cells, where the average number of particles per cell varied between $1$ and $50$, and the history length $k$ was sampled
uniformly between $0$ and $10$. The neural network architecture uses two separate branches depending on $k$. A DeepONet-based architecture~\cite{Lu_2021} was used for $k=0$, whereas a Transformer-based architecture~\cite{vaswani2023attentionneed} was used for $k>0$.    

\paragraph{Statistical reflection symmetry}
Symmetry of particle current dictates that the PDF of $J(\bm{x}, t, N_{L}=a, N_{R}=b)$ is the same as the PDF of $-J(\bm{x}, t, N_{L}=b, N_{R}=a)$.
In the FM model this implies that
\begin{equation}
    u_{\tau}^{\theta}(Z_{\tau}, \bm{N}_{L}^{k}, \bm{N}_{R}^{k}, \bm{J}^{k}) = - u_{\tau}^{\theta}(-Z_{\tau}, \mathcal{R}(\bm{N}_{L}^{k}), \mathcal{R}(\bm{N}_{R}^{k}), -\bm{J}^{k}),
\end{equation}
where $\mathcal{R}(\bm{N}_{L}^{k}) = \bm{N}_{R}^{k}$, and $\mathcal{R}(\bm{N}_{R}^{k}) = \bm{N}_{L}^{k}$. This property was enforced by reconstructing $u_{\tau}^{\theta}$ as:
\begin{equation}
    u_{\tau}^{\theta}(Z_{\tau}, \bm{N}_{L}^{k}, \bm{N}_{R}^{k}, \bm{J}^{k}) = 0.5 \left[v_{\tau}^{\theta}(Z_{\tau}, \bm{N}_{L}^{k}, \bm{N}_{R}^{k}, \bm{J}^{k}) - v_{\tau}^{\theta}\left(-Z_{\tau}, \mathcal{R}(\bm{N}_{L}^{k}), \mathcal{R}(\bm{N}_{R}^{k}), -\bm{J}^{k}\right)\right],
\end{equation}
where $v_{\tau}^{\theta}$ has the optimizable parameters.
\section{Results}
\label{others}
We tested our numerical method on a variant of the Kramers first passage time problem~\cite{kramers1940brownian}.  We simulated $5120$ independent realizations of a one-dimensional system consisting of $100$ finite-volume cells of size $\Delta x = 1 \times 10^{-2}$. The time step for evolving both the SPDE and the Brownian particle simulation was chosen as $\Delta t = 3 \times 10^{-6}$ with homogeneous Dirichlet boundary conditions at the two ends of the domain. The dynamics includes an external potential of the form $V(x) = (x - \alpha)^{2} (x-\beta)^{2}$, where $\alpha=0.3$ and $\beta = 0.7$. The external potential has minima at $x=0.3$ and $x=0.7$, and a local maximum at $x=0.5$. The system was initialized with $10^{3}$ total particles that were distributed in the region from $x=0.2$ to $x=0.4$ based on an equilibrium distribution under the external potential, $\frac{1}{\mathcal{Z}} \exp({-2\mu V(x)})$, where $\mathcal{Z}$ is the normalization constant and $\mu = 2 \times 10^{3}$. The primary objective in this numerical experiment was to track the rate at which particles transition from the region around $x=\alpha$ into the region around $x=\beta$, across the energy barrier of height $V_b$ at $x=0.5$. We computed the particle statistics in the target region around $x=\beta$ where $V(x) \leq \frac12 V_b$.

Each system was simulated for $\sim 3\times10^3$ steps using the three models: Brownian random-walkers, regularized DK model, and the FM model where the SPDE flux $\bm{F}^{B}$ is generated by the FM method. To emphasize non-Markovian effects, the simulations were run with the FM models both with a history length of $k=10$, and the Markovian limit without history, i.e., $k=0$. In the FM method, a uniform 20 step Euler time discretization was used in the ODE solver for sampling the target distribution.  All the simulations were performed on CPUs, and the most expensive non-Markovian FM model scaled well up to $100$ cells per CPU, and averaging 9 seconds per time step. 

\begin{figure}
    \centering
    \includegraphics[width=\textwidth]{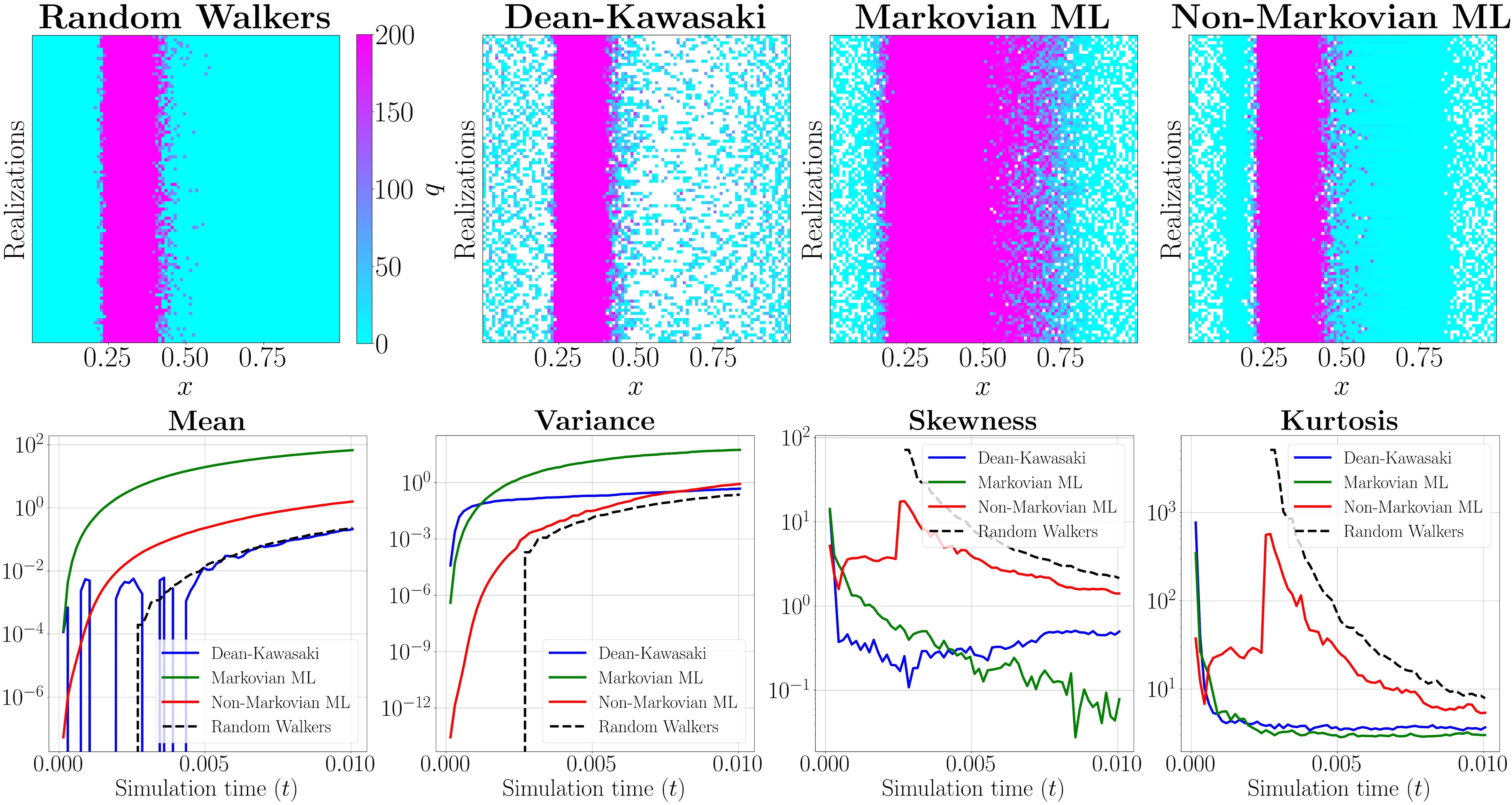}
    \caption{
    The top row shows a subset of 100 independent one-dimensional realizations (along vertical direction) at time $t = 5.1\times 10^{-3}$ simulated using Brownian random walkers, the regularized Dean–Kawasaki (DK) model, and the two flow matching (FM) models: Markovian and non-Markovian. The white patches denote negative values in the simulation. The bottom row compares the time evolution of the statistics of particle count in the target region around $x=0.7$.
    }
    \label{fig:double_well}
\end{figure}

Figure~\ref{fig:double_well} depicts independent realizations of an early-time state of the system. The regularized DK model results in high occurrence of nonphysical negative densities as compared to the ML models. While the DK model well-predicts mean particle densities in the target region, the non-Markovian ML model more accurately captures the spatial particle distribution and also the higher-order statistical moments (see bottom row), thus demonstrating significant non-Markovian effects at early times.

\textbf{Code and data availability:} The code and data are open-source and available at the following links: \href{https://github.com/siddanib/PAI26_Submission_76}{code} and \href{https://huggingface.co/datasets/bsiddani/PAI26_Submission_76}{data}.

\section{Conclusion}
In this work, we used a generative flow matching framework to model non-Gaussian and non-Markovian effects (i.e., memory) in coarse-grained stochastic partial differential equations that represent stochastic particle systems. Using non-interacting Brownian particles as an example, we showed that the proposed approach more accurately captures short-time dynamics than Markovian models such as the DK equation. Furthermore, our flow matching based method reproduced higher-order statistics reasonably well at early times of the system evolution. However, at later times, the model predictions began to deviate significantly from the particle simulations, with the error increasing with time. While the current methodology is more computationally expensive than non-interacting Brownian dynamics simulations, it is expected to be significantly more efficient for stochastic particle systems with complex, long-ranged interactions, where costly neighbor-list construction and time step constraints incur significant computational costs.

\begin{ack}
This work was supported by the LDRD program of Lawrence Berkeley National Laboratory under U.S. DOE Contract No. DE-AC02-05CH11231, and by the U.S. DOE, Office of Science (SC), Office of ASCR, Applied Mathematics Program under Contract No. DE-AC02-05CH11231. This research used the resources of the NERSC, a DOE-SC User Facility, under Contract No. DE-AC02-05CH11231.
\end{ack}

{
\small
\bibliographystyle{unsrt}
\bibliography{references,Literature}
}




\end{document}